# Sit-to-Stand Transitions Detection and Duration Measurement Using Smart Lacelock Sensor

Md R. Islam, Md R. Haque, Elizabeth Choma, Shannon Hayes, Siobhan McMahon,
Xiangrong Shen, and Edward Sazonov, *Senior Member*, *IEEE*

**Abstract**—Postural stability during movement is fundamental to independent living, fall prevention, and overall health, particularly among older adults who experience age-related declines in balance, muscle strength, and mobility. Among daily functional activities, the Sit-to-Stand (SiSt) transition is a vital indicator of lower-limb strength, musculoskeletal health, and fall risk, making it an essential parameter for assessing functional capacity and monitoring physical decline in aging populations. This study presents a methodology SiSt transition detection and duration measurement using the Smart Lacelock sensor, a lightweight, shoe-mounted device that integrates a load cell, accelerometer, and gyroscope for motion analysis. The methodology was evaluated in 16 older adults (age: 76.84 ± 3.45 years) performing SiSt tasks within the Short Physical Performance Battery (SPPB) protocol. Features extracted from multimodal signals were used to train and evaluate four machine learning classifiers using a 4-fold participant-independent cross-validation to classify SiSt transitions and measure their duration. The bagged tree classifier achieved an accuracy of 0.98 and an F1 score of 0.8 in classifying SiSt transition. The mean absolute error in duration measurement of the correctly classified transitions was 0.047 ± 0.07 seconds. These findings highlight the potential of the Smart Lacelock sensor for real-world fall-risk assessment and mobility monitoring in older adults.

*Index Terms*—Sit-to-Stand Transition, Fall Risk, Load cell, Inertial Measurement Unit (IMU), Machine Learning, Mobility Monitoring, Wearable Sensors

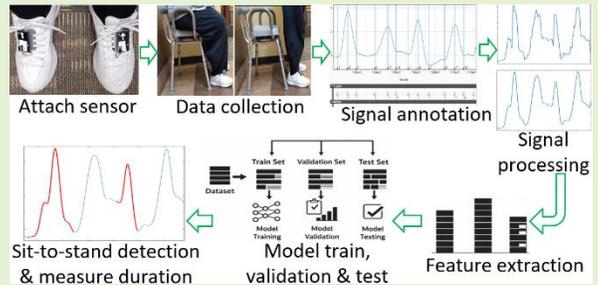

Attach sensor    Data collection    Signal annotation    Signal processing    Sit-to-stand detection & measure duration    Model train, validation & test    Feature extraction

## I. Introduction

THE ability to maintain balance during movement is critical for preserving independence, minimizing fall risk, and promoting overall well-being, particularly among older adults, for whom declines in muscle strength, balance, and reaction time substantially increase fall susceptibility [1], [2]. Falls are a leading cause of injury, hospitalization, and loss of independence in adults over 65 years, and approximately one in four older adults experiences a fall each year [1]. Among daily functional activities, the Sit-to-Stand (SiSt) transition is a fundamental movement that serves as a sensitive marker of neuromuscular performance, postural control, and lower-limb strength in aging populations [2], [3], [4]. Difficulties in executing this transition often indicate early mobility impairments and an elevated risk of recurrent falls [4], [5]. Therefore, reliable detection and continuous monitoring of SiSt transitions in older adults hold significant potential for early diagnosis, personalized rehabilitation, and long-term mobility assessment [6], [7].

Conventional clinical tests, such as the Five-Times-Sit-to-Stand (5xSiSt) and Thirty-Second Chair Stand Tests [6], [7], provide standardized benchmarks for evaluating lower-limb strength and endurance but are typically restricted to controlled settings and offer only episodic observations of performance. They fail to capture the dynamic variability of SiSt execution in natural environments, where subtle compensatory movements or fatigue-related changes often manifest [8]. Continuous, real-world monitoring could therefore provide a richer and more ecologically valid understanding of mobility, balance, and fatigue in older adults—key predictors of fall risk and functional decline [8].

Recent studies have examined a range of techniques for classifying SiSt transitions, each accompanied by specific limitations. Methods based on pressure sensors [9] [10] or motion data from inertial measurement units (IMUs) [11], [12], [13], [14], [15], [16], [17], [18] have demonstrated potential but often suffer from reduced accuracy in uncontrolled or outdoor environments and may compromise user comfort due to sensor placement. Similarly, approaches involving ground reaction

†This work was supported by the National Science Foundation under award 1734501. (Corresponding author: Edward Sazonov)
Md R. Islam and Edward Sazonov are with the Department of Electrical and Computer Engineering, The University of Alabama, Tuscaloosa, AL 35487 (e-mail: mislam24@crimson.ua.edu; esazonov@eng.ua.edu)
Md R. Haque is with Genesis Medtech USA Inc (e-mail: rejwanul.haque@gmedtech.com)

Elizabeth Choma is with Whitworth University's Doctor of Physical Therapy program (e-mail: lchoma@whitworth.edu)
Xiangrong Shen is with the Department of Mechanical Engineering, The University of Alabama, Tuscaloosa, AL 35487 USA (e-mail: xshen@eng.ua.edu).
Shannon Hayes and Siobhan McMahon is with the University of Minnesota School of Nursing (e-mail: shannonh6200@gmail.com; skmcmaho@umn.edu)

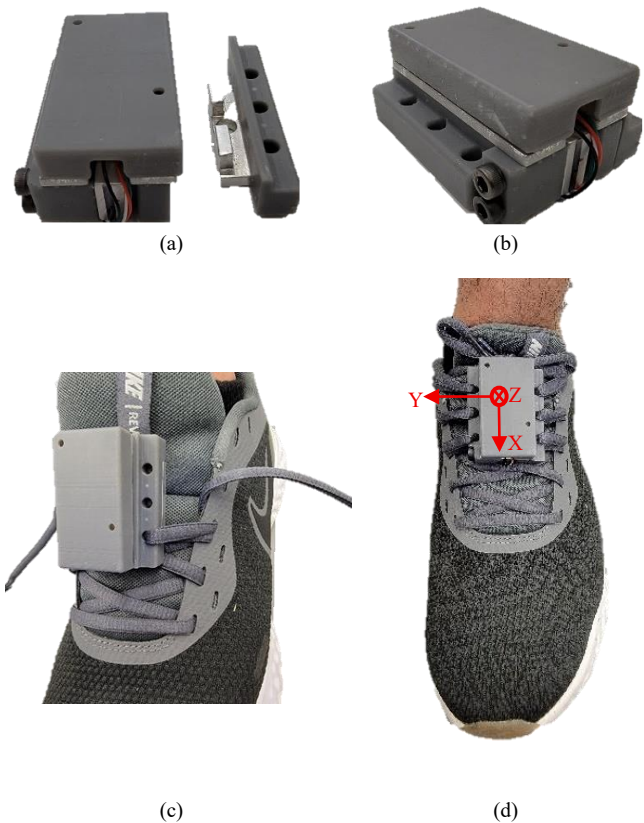

Fig. 1. (a) Smart Lacelock sensor assembly, (b) Assembled sensor, (c) Attaching the sensor to the shoe, (d) Completed sensor attachment on the shoe and the orientation of the IMU

force estimation or other body-worn sensors are frequently challenged by issues such as wearer discomfort, sensor reliability, and the computational complexity of the algorithms required for real-time processing [12], [19].

Other approaches, such as embedding sensors within bionic knee exoskeletons [17] or utilizing electromyography (EMG) signals [20], offer valuable physiological and biomechanical insights. However, these techniques are often constrained by their invasive nature, sensitivity to parameter tuning, and substantial computational requirements, which may limit their practicality in real-world settings [11], [13], [15], [21].

Sensors embedded into shoes and insoles [22], [23], [24] allow to measure the pressure under the foot and inertial forces acting on the foot in a comfortable form, however, these sensors typically have significant durability limitations due to the high forces acting on the electronics during ambulation, and create challenges in the recharging the battery.

To address these limitations, this study uses the Smart Lacelock sensor, a shoe-mounted sensor integrating a load cell and an IMU, to enable multimodal measurement of shoelace tension and foot motion (Fig. 1). This unique configuration provides biomechanical information analogous to insole-based systems while improving durability, ease of recharging, and user comfort—attributes particularly important for older adults. Its lightweight, unobtrusive design eliminates the need for invasive or body-worn sensors and facilitates effortless daily use, even by individuals with limited dexterity or balance.

By fusing load cell and IMU signals within a machine-learning framework, the sensor enables automated detection of SiSt transitions during natural activity, overcoming the limitations of traditional clinic-based assessments. This study presents the development and validation of the Smart Lacelock sensor-based methodology for detecting sit-to-stand transitions and quantifying transition duration. The following sections describe the system design, experimental protocol, validation metrics, and performance results.

## II. GUIDELINES FOR MANUSCRIPT PREPARATION

### A. Abbreviations and Acronyms

The Smart Lacelock sensor combines a 9-degrees-of-freedom (DOF) inertial measurement unit (IMU) for three-dimensional motion tracking with a miniature load cell for real-time shoelace tension measurement [25]. As shown in Fig. 1(a), the system comprises two assemblies, with the fully assembled configuration depicted in Fig. 1(b). In practice, only the assembled sensor is used. The sensor is mounted by threading the shoelaces sequentially through the shoe's eyelets from bottom to top, as illustrated in Fig. 1(c). Once positioned, the shoelace is tightened to ensure a snug but comfortable fit, and the sensor is locked for data acquisition (Fig. 1(d)).

The integrated electronics includes a Straight Bar TAL221 load cell, an MPU-9250 IMU (containing a 3-axis accelerometer and a 3-axis gyroscope), and an STM32L476RG Cortex-M4 ultra-low-power microcontroller unit (MCU). Signals are recorded at 512 Hz and stored on a 32 GB micro-SD card. The system also incorporates a HX711 24-bit ADC for precise load cell signal conversion and a micro-USB interface for data retrieval, battery charging, and MCU timestamp synchronization.

Motion sensing is performed using the IMU's ±16 g accelerometer and ±2000°/s gyroscope, each with 16-bit resolution, ensuring high-fidelity motion capture.

### B. Data Collection and Signal Annotation

The University of Minnesota Institutional Review Board approved all study procedures, and informed consent was obtained from all participants. This study recruited 16 older adults (3 males and 13 females), aged 71–83 years (mean age: 76.84 ± 3.45 years). The Smart Lacelock sensor was securely attached to each participant's shoe by trained research personnel prior to administering the Short Physical Performance Battery (SPPB) [6] test.

The SPPB included three subtests: (1) static balance, where participants maintained three progressively challenging stances for 10 seconds each; (2) gait speed, measured over a 4-meter walk; and (3) chair stand, which assessed the time required to complete five consecutive SiSt repetitions with arms folded across the chest.

The experiments were recorded with a smartphone camera. Prior to each trial, the internal clock of the Smart Lacelock sensor was synchronized with an NTP (Network Time Protocol) server using the NetTime application. The smartphone was connected to the internet to ensure its system clock was up to date. At the beginning of each experiment, a light tap by hand was applied to each sensor to generate a distinct peak in the load cell signal. This tapping event was simultaneously recorded using video. During signal processing, the timestamp corresponding to the peak in the load cell signal was aligned with the corresponding tapping moment observed in the video recording. Any detected temporal offset was compensated by adjusting the sensor timestamps accordingly (i.e., by adding or subtracting the measured delay). This procedure minimized time drift and ensured close synchronization between the sensor and the smartphone clocks at the start of data collection. At the conclusion of each experiment, the tapping procedure was repeated to assess clock drift between the sensor and the smartphone. The elapsed time measured by the sensor clock during the experiment was compared with the elapsed time recorded by the smartphone clock, and among all 16 experiments, the observed maximum drift was 0.003s.

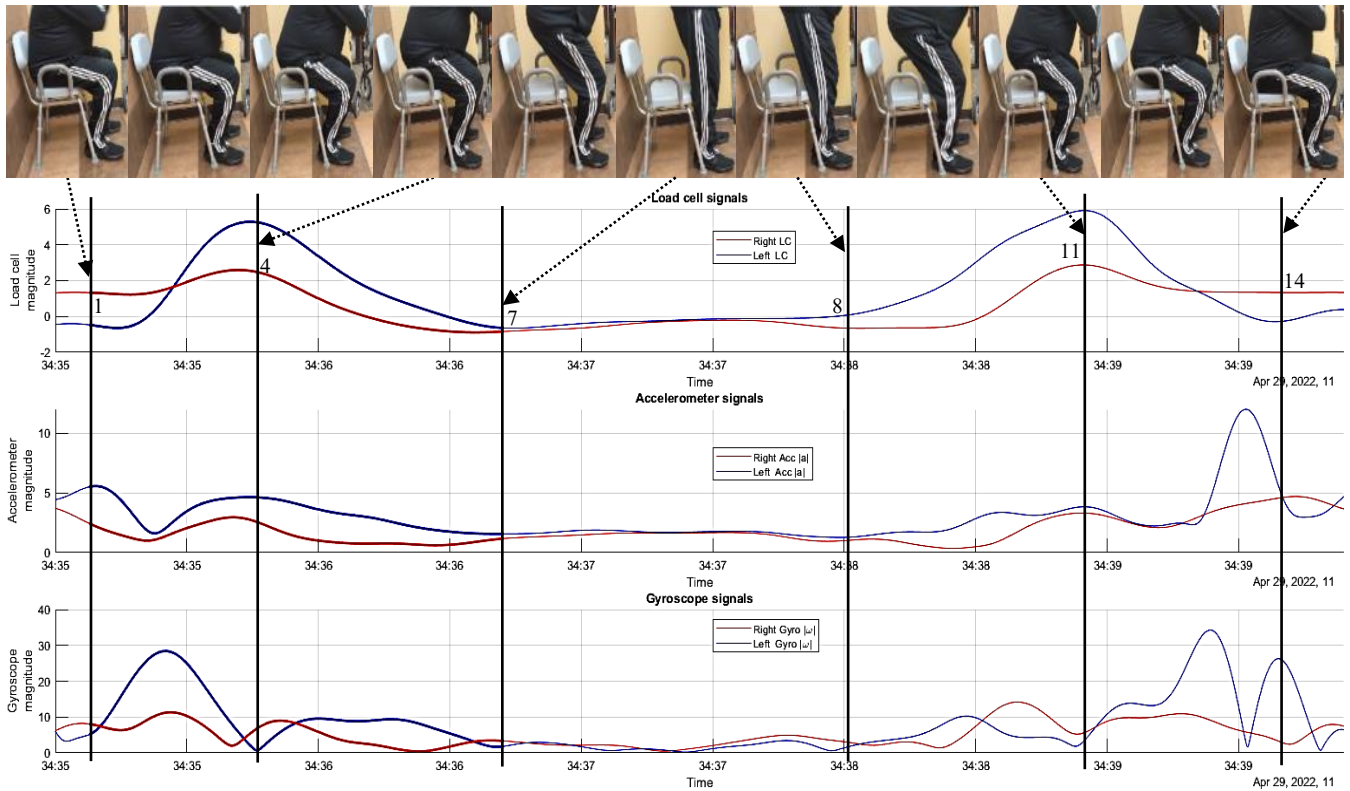

Fig. 2. Annotation of Sit-to-Stand and Stand-to-Sit transitions. 1 = initiation of forward trunk lean during rising, 4 = maximal forward lean just before vertical ascent, 7 = final backward lean during recovery, 8 = initiation of forward lean to begin sitting, 11 = point of maximal forward lean, 14 = final backward lean as the subject regains a stable seated posture. Markers 1 through 7 is the SiSt transition segment, markers 8 through 14 is the StSi transition. Right and left mean signals from the right and left feet, respectively. In the bottom two plots, the Accelerometer magnitude was computed as the Euclidean norm of the tri-axial accelerometer signals using (1), and the Gyroscope magnitude was computed as the Euclidean norm of the tri-axial Gyroscope signals using (1)

SiSt transitions were annotated by cross-referencing the video recordings with sensor data. Transition events were visually identified in the video and annotated in the load cell signal, using MATLAB's Signal Labeler application. The load cell signal exhibited a characteristic sagittal displacement pattern consistent with previously reported observations [26]. Specifically, as described in Fig. 2, point 1 marks the initiation of forward trunk lean during SiSt, while point 4 denotes the point of maximal forward lean just before vertical ascent. Point 7 represents the final backward lean during recovery, completing the rising phase. During the stand-to-sit (StSi) transition, point 8 indicates the initiation of forward lean to begin sitting, point 11 captures the point of maximal forward lean in this phase, and point 14 marks the final backward lean as the subject regains a stable seated posture. Fig. 2 presents a representative example of the annotation process, illustrating a load cell signal. Because all sensor modalities shared a common timestamp (x-axis), the marked SiSt start and end timestamps from the load cell signal were then copied to the corresponding accelerometer and gyroscope timestamps to ensure consistent labeling across all sensor modalities. It is important to note that, when annotating, both foot signals were labeled synchronously at the same start and end time points, ensuring temporal alignment.

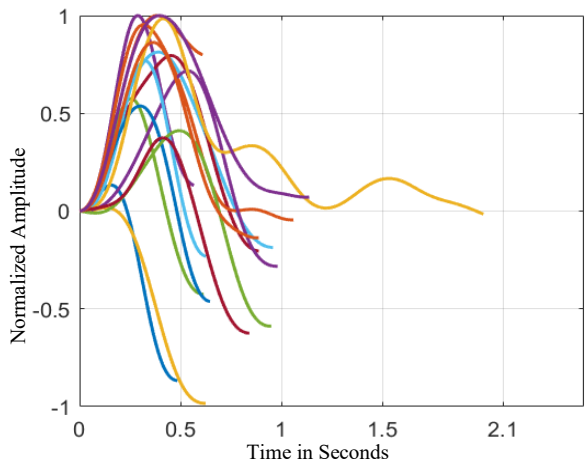

Fig. 3. Shortest SiSt transitions of all participants (Left foot Load cell signal)

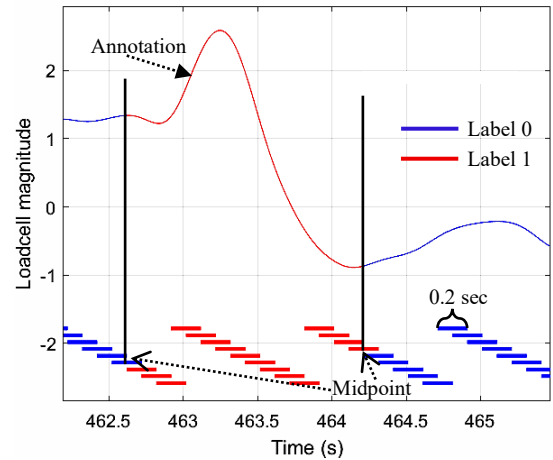

Fig. 4. Visualization of signal labeling

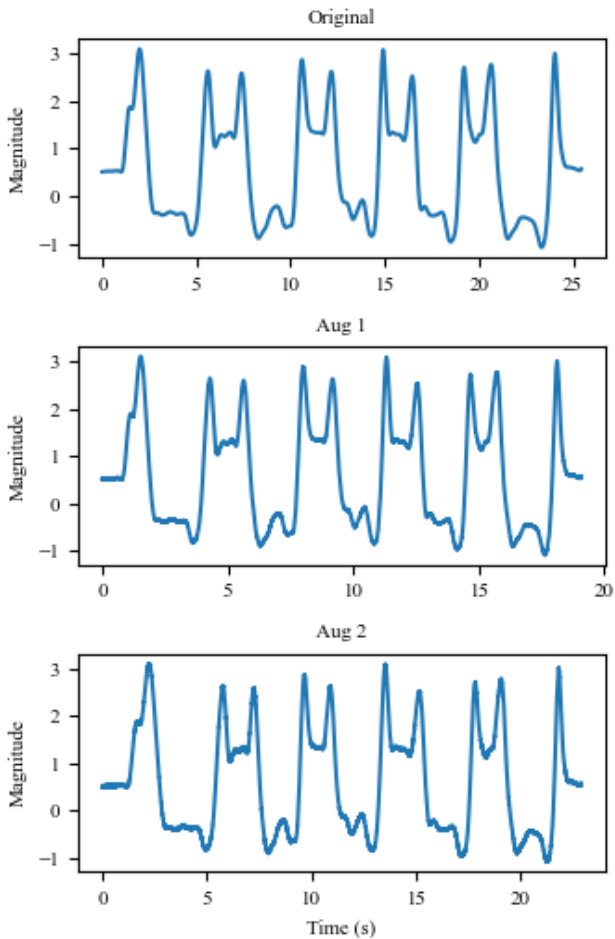

Fig. 5. Original and augmented load cell signal

### C. Signal Labeling

This section describes the procedure for converting the annotations into classification labels. The duration of the SiSt transitions varied across participants. The mean SiSt transition duration was 1.52s, with a standard deviation of 0.36s. The shortest SiSt transition was 0.39s, and the longest was 2.87s. Fig. 3 illustrates the shortest SiSt transitions among the participants. To account for inter-individual variability in SiSt transition duration and to determine the optimal window size for classification, several window lengths were tested. Seven base sliding window lengths, i.e., 0.1s, 0.2s, 0.3s, 0.4s, 0.5s, 0.6s, and 0.7s were defined and are henceforth referred to as primary windows. These primary windows served as building blocks for generating a series of extended windows with increasing durations, created by concatenating an odd number (n) of consecutive primary windows until a window length at which the classifiers' performance began to decline.

For instance, with a primary sliding window of 0.1s, the extended sliding windows included 0.3 s, 0.5 s, 0.7 s, …, and so on (i.e., 0.1s*n where n = 3, 5, 7, …). Similarly, for a . primary window of 0.2 s, extended sliding windows of 0.6 s, 1.0 s, 1.4 s, and 1.8 s were formed by concatenating n = 3,5, 7, and 9 consecutive segments. So, the window size with n=1 refers to the primary window.

Signal labeling was performed using a sliding-window approach with a stride equal to the primary window size. Each window was assigned a binary label based on whether its midpoint fell within the annotated SiSt transition (labeled 1) or outside it (labeled 0). Fig. 4 illustrates this labeling process for a 0.2s window.

### D. Signal Augmentation

To mitigate the class imbalance between SiSt transitions and other activity data (initially, the SiSt-to-other activity ratio was 5:95), a twofold data augmentation strategy was implemented. This approach targeted both the raw SiSt signals and their corresponding extracted features.

For signal-level augmentation, the Time Series Augmentation (TSAUG) framework [27] was applied to the training folds, incorporating a set of time-series transformations, including noise injection, time warping, and frequency-domain augmentations such as Fast Fourier Transform (FFT) phase and amplitude perturbation and Discrete Cosine Transform (DCT) phase perturbation. All augmentation parameters were constrained to remain within physiologically plausible limits to preserve the natural characteristics of SiSt dynamics. For instance, for time-warping augmentation, the upper and lower limits were randomly selected from a uniform distribution where the distribution's lower and upper limits were defined by the minimum and maximum SiSt transition durations observed across participants. This ensured that the augmented segments remained physiologically realistic and representative of natural variability. An illustration of the TSAUG-augmented SiSt signals is provided in Fig. 5.

After the signal level augmentation, the SiSt to other class ratio was about 45:55. Then, the feature-level augmentation was conducted on the training folds using the Synthetic Minority Over-sampling Technique (SMOTE) [28]. This approach synthetically increased the number of feature instances corresponding to the SiSt class, aligning their count with that of more frequently occurring activity segments. Following augmentations, the class distribution improved substantially, resulting in a more balanced 50:50 ratio of SiSt transition to other activity data.

### E. Feature Extraction

Prior to feature extraction, all sensor signals were standardized using z-normalization to ensure uniform scaling across modalities. From each signal window, 19 temporal, 16 statistical, and 12 frequency-domain features were extracted. The details of the features are provided in the supplemental

TABLE I.  MOST FREQUENTLY SELECTED FEATURES IN TRAINING

| Sl No. | Description | Sl No. | Description |
|---|---|---|---|
| 1. | Mean load magnitude difference ($|\overline{L\_right} - \overline{L\_left}|$) | 11. | Slope sign change *(GZl)* |
| 2. | Mean acceleration magnitude *(Acc_right)* | 12. | Gyro magnitude correlation *(Gyro_right*Gyro_left)* |
| 3. | Load cell signal correlation *(L_right*L_left)* | 13. | Mean absolute value *(L_left)* |
| 4. | Average absolute difference *(AZ_right)* | 14. | RMS *(AZ_left)* |
| 5. | Slope sign change *(L_right)* | 15. | Time to peak *(L_right)* |
| 6. | Mean Accelerometer magnitude product *(Acc_right*Acc_left)* | 16. | Max to min difference *(L_right)* |
| 7. | Mean time between peaks *(AZ_left)* | 17. | Vertical acceleration correlation *(AZ_right*AZ_left)* |
| 8. | RMS *(L_right)* | 18. | Mean to max ratio *(L_left)* |
| 9. | Spectral power *(AZ_right)* | 19. | Spectral Entropy *(GZ_right)* |
| 10. | Max to RMS Ratio *(L_right)* | 20. | RMS mean ratio *(AZ_left)* |

Here, L is load cell signal, AZ is Z-axis acceleration signal, GZ is Z-axis gyroscope signal, _right means signal from right foot, _left means signal from left foot. Acceleration (Acc_right or Acc_left) and gyroscope magnitude (Gyro_right or Gyro_left) was computed using (1).

document. The acceleration and gyroscope magnitude stated in Table I were computed using the following equation,

$$sig\_mag = \sqrt{sig_x^2 + sig_y^2 + sig_z^2} \qquad (1)$$

Here, $sig\_mag$ means signal magnitude, $sig_x$ means signal from x-axis of the sensor, $sig_y$ means signal from y-axis sensor, $sig_z$ means signal from z-axis sensor. The orientation is illustrated in Fig. 1(d).

In addition to conventional statistical, time-domain, and frequency-domain features, correlation between left and right foot signals was also extracted. That is because, as mentioned earlier, the signals are symmetrical in left and right foot during SiSt transitions, while they are mostly asymmetrical in other activities. Six transition-based features were extracted. For each foot, transition-based features included changes in z-axis acceleration, time-to-peak of z-axis acceleration, and time-to-peak of the load cell signal. These features are particularly important because SiSt is defined by short, transient phases with distinctive signal transitions.

From each sliding window, a total of 667 features are extracted from 14 signals, consisting of seven signals per foot (six IMU signals and one load cell signal).

### F. Training and Testing Classifiers:

In this study, four machine learning classifiers—Boosted Tree, Bagged Tree, Medium Gaussian Support Vector Machine (SVM), and Weighted K-Nearest Neighbors (KNN), were employed for the classification of SiSt transitions using load cell and IMU signals [29], [30], [31], [32]. These classifiers were selected based on their demonstrated effectiveness in handling nonlinear, high-dimensional, and sensor-based time-series data commonly observed in human movement studies. The classifiers' performance was also evaluated without load cell signal features to determine whether load cell signal contributes to improving the SiSt classification performance. Classifier optimization was conducted using a grid search on the training folds to identify the most effective hyperparameter configurations.

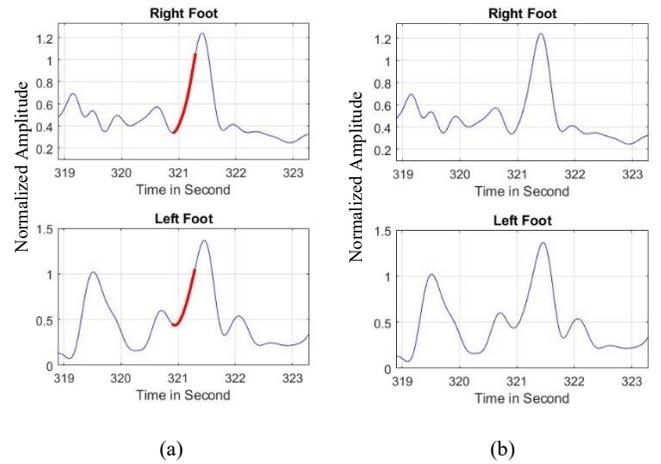

Fig. 6. (a) Initially detected Sit-to-Stand segment. (b) Discarded detection with majority voting.

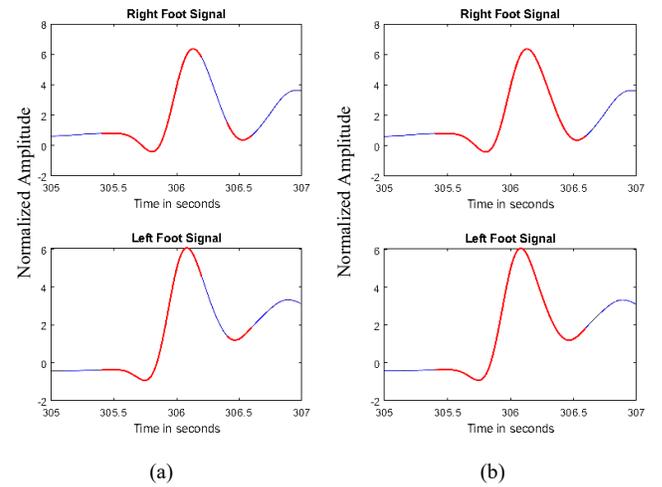

Fig. 7. (a) Before applying majority voting; (b) After applying majority voting

A 4-fold internal cross-validation split by participants was employed, so that data from the same participant belonged only to one of the folds. Classifiers were trained on three folds (12 participants) and evaluated on the remaining fold (4 participants) in each iteration. This approach ensured independent validation and prevented any overlap between training and testing data.

To select the most informative features in each sliding window from the initial pool of 667 features, the Minimum Redundancy Maximum Relevance (mRMR) algorithm was applied to the training dataset during each cross-validation iteration [33]. Feature relevance scores were ranked in descending order to prioritize the most discriminative features. F1-score was used as the primary performance metric because the dataset was highly imbalanced. Unlike accuracy, F1 balances precision and recall, providing a more reliable evaluation of model performance when one class significantly outnumbers the other. F1 score did not increase beyond 20 features, leading to the selection of the top 20 features for final model development. The most frequently selected features across all training folds are listed in Table I.

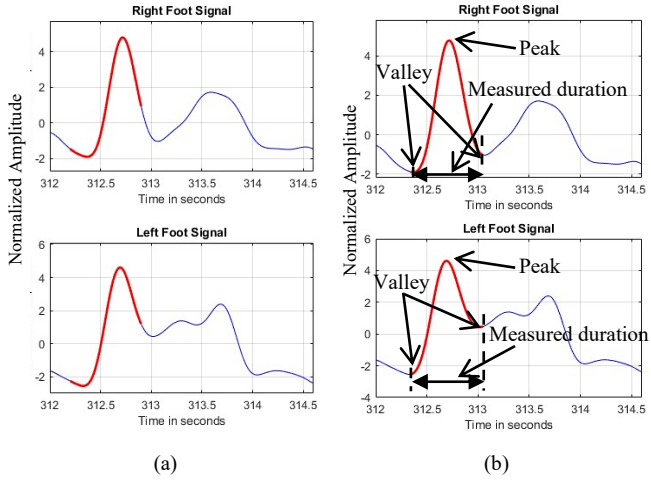

Fig. 8. (a) Detected SiSt segment in red line. (b) Measured SiSt duration from valley to peak to valley in red line.

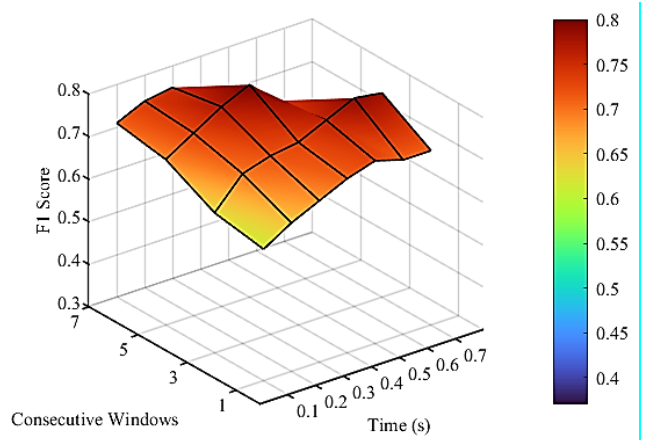

Fig. 9. Performance of the bagged tree classifier for different sliding window sizes.

### G. Post Processing:

In the post-processing stage, a 5-window majority voting temporal smoothing step was applied to enforce state continuity. The number of windows used in majority voting was selected through grid search optimization on training folds. Fig. 6 and 7 illustrate the majority voting scheme.

### H. SiSt Duration Measurement

As shown in Fig. 1 and Fig. 2, the SiSt transition in the load cell signal begins at a valley corresponding to the initiation of forward trunk lean, rises to a peak representing the maximum forward lean before vertical ascent, and then returns to a valley associated with backward lean during recovery. Accordingly, for each detected SiSt segment, the transition duration was measured as the interval from the initial valley to the peak and back to the final valley. After preprocessing, the initial and final indices of each SiSt segment were first identified. When multiple consecutive SiSt transitions were detected, the initial index was set to the first sample of the first segment, and the final index was set to the last sample of the last segment. Within this selected index range, the peak was defined as the maximum signal magnitude. The initial valley was selected as the minimum value around the first classified SiSt window of that transition to the left of the peak, and the final valley as the minimum value near the last classified SiSt window of that transition to the right of the peak. This procedure is illustrated in Fig. 8.

### I. Performance Evaluation

In evaluating the classification outcomes, four key metrics were employed to comprehensively assess the performance of the sensor in detecting SiSt transitions. The classification results were categorized into true positives, true negatives, false positives, and false negatives. True Positives (TP) represent windows correctly classified as SiSt transitions. True Negatives (TN) encompass windows accurately classified as non-transition windows without any SiSt transition. False Positives (FP) signify windows incorrectly classified as SiSt transition windows. False Negatives (FN) include SiSt transition windows that were not identified as such. To quantitatively evaluate the classifier's performance, four metrics were computed. These metrics include accuracy, precision, recall, and F1 score, which were calculated using the following formulas:

$$Accuracy = \frac{TP + TN}{TP + TN + FP + FN} \quad (2)$$

$$Precision = \frac{TP}{TP + FP} \quad (3)$$

$$Recall = \frac{TP}{TP + FN} \quad (4)$$

$$F1\ Score = \frac{2 * Precision * Recall}{Precision + Recall} \quad (5)$$

The ground-truth SiSt duration was obtained from annotated SiSt transitions. The accuracy of the SiSt duration measurements was evaluated using root mean square error (RMSE) and mean absolute error (MAE). The corresponding equations are provided below.

$$RMSE = \sqrt{\frac{\sum_{i=1}^{n}(y_i - \hat{y}_i)^2}{n}} \quad (6)$$

$$MAE = \frac{1}{n}\sum_{i=1}^{n}|y_i - \hat{y}_i| \quad (7)$$

Here, $y_i$ is the actual SiSt duration, $\hat{y}_i$ is the measured SiSt duration.

### III. RESULTS

The analysis of variance test of the four classifiers' performance across different sliding window lengths shows no statistically significant variation (p > 0.05). Therefore, the subsequent sections present results from the Bagged Tree classifier (NumTrees = 200, MinLeafSize = 5), which was selected for detailed analysis of SiSt classification because it attained the highest F1 score among other classifiers.

Fig. 9 shows the SiSt transition classification performance of the Bagged Tree classifier for different sliding window sizes.

TABLE II. CLASSIFICATION PERFORMANCE WITH AND WITHOUT USING
LOAD CELL (SHOELACE TENSION) DATA

| Sliding Window | F1 Score | |
|---|---|---|
| | *With load cell features* | *Without load cell features* |
| 0.1s*n | 0.69 ± 0.05 | 0.50 ± 0.07 |
| 0.2s*n | 0.73 ± 0.06 | 0.56 ± 0.06 |
| 0.3s*n | 0.77 ± 0.04 | 0.57 ± 0.06 |
| 0.4s*n | 0.80 ± 0.07 | 0.62 ± 0.05 |
| 0.5s*n | 0.74 ± 0.08 | 0.58 ± 0.06 |
| 0.6s*n | 0.61 ± 0.06 | 0.55 ± 0.06 |
| 0.7s*n | 0.51 ± 0.09 | 0.47 ± 0.07 |

Here, n=5

As previously noted, due to class imbalance, the F1-score was chosen to evaluate performance since it balances precision and recall, providing a more dependable assessment when one class is predominant. The time axis represents the primary window durations (0.1 s, 0.2 s, 0.3 s, 0.4 s, and so on), while the consecutive windows axis indicates the number of primary windows consecutively added to form an extended window. The highest F1 score of 0.8 was achieved using 5 consecutive 0.4s windows, corresponding to a total window length of (0.4s*5), or 2s. Consequently, adding more windows beyond this point (e.g., 7 windows) to increase the sliding window size reduced classification performance, as indicated by the decline in the plot. Fig. 10-11 shows the SiSt transition classification performance using a 0.4s*5 sliding window (2s). In these figures, the blue lines represent the detected non-SiSt activity, while the red lines illustrate the detected SiSt transitions.

Table II presents the mean and standard deviation of the F1 score across folds of the bagged tree classifier, obtained by including and excluding the load cell signal features to evaluate its contribution to SiSt detection performance. Inclusion of load cell data resulted in a statistically significant improvement in F1-score compared to the IMU-only model, as determined by a

TABLE III. SISt DURATION MEASUREMENT PERFORMANCE

| Sliding Window | Duration Error for True Positives (s) | | Duration Error for All Detections (s) | |
|---|---|---|---|---|
| | *MAE* | *RMSE* | *MAE* | *RMSE* |
| 0.1s*n | 0.261±0.15 | 0.372±0.19 | 0.599±0.17 | 0.725±0.28 |
| 0.2s*n | 0.191±0.09 | 0.293±0.13 | 0.448±0.20 | 0.552±0.27 |
| 0.3s*n | 0.086±0.11 | 0.162±0.17 | 0.329±0.15 | 0.408±0.24 |
| 0.4s*n | 0.047 ±0.07 | 0.079±0.12 | 0.238±0.12 | 0.362±0.19 |
| 0.5s*n | 0.091±0.1 | 0.116±0.14 | 0.268±0.88 | 0.364±0.22 |
| 0.6s*n | 0.174±0.12 | 0.247±0.20 | 0.289±0.13 | 0.387±0.24 |
| 0.7s*n | 0.325±0.18 | 0.399±0.22 | 0.382±0.16 | 0.459±0.28 |

Here, n=5

paired t-test (p < 0.05), confirming its contribution to transition detection.

As the 2s sliding window (0.4s*5) achieved the best classification performance and was therefore selected for reporting the SiSt transition duration measurement performance. The ground truth for calculating the measurement was the SiSt annotations on the load cell signal mentioned earlier.

Two sets of duration error evaluation metrics were reported: "Duration Error for True Positives" and "Duration Error for All Detections", expressed using RMSE and MAE with standard deviation across folds. For determining the "Duration Error for True Positives", errors were computed only for predicted SiSt segments that had equal or more than 50% overlap with a ground-truth SiSt event, inspired by the Intersection over Union (IoU) framework widely used in detection problems. This criterion ensures that a detected segment captures the majority of the true transition duration while maintaining tolerance to minor boundary misalignments inherent to sliding-window detection methods. The "Duration Error for All Detections" accounted for incorrectly classified transitions, i,e. FP were assigned an error equal to that predicted segment's duration.

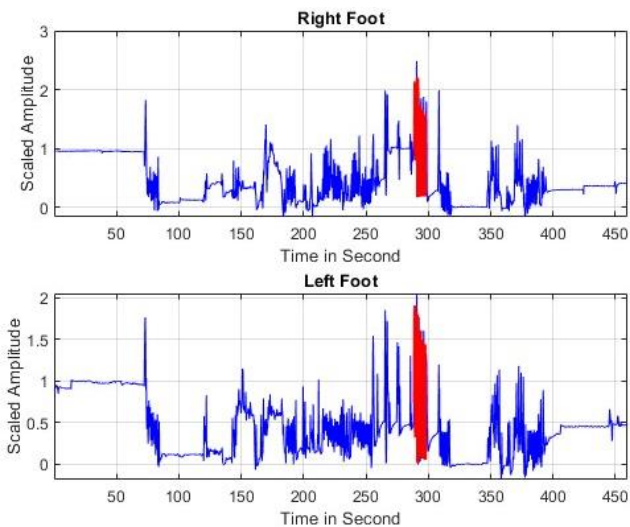

Fig. 10. Classifier performance on SiSt transition detection. It illustrates the performance on the whole signal for 0.4s*n sliding window (n=5). Red lines represent SiSt, blue lines represent other activities.

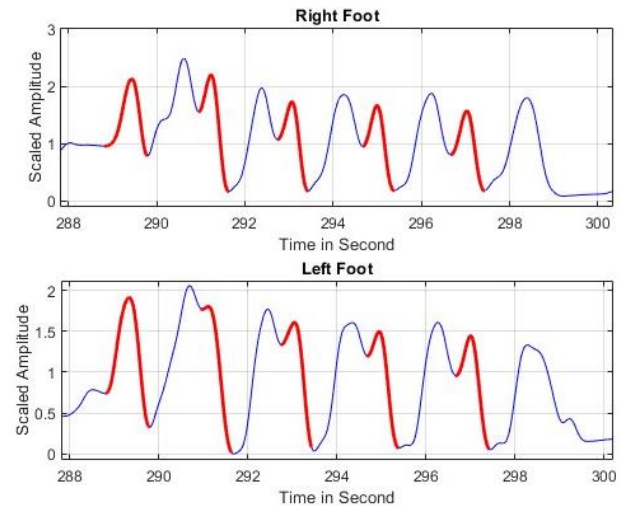

Fig. 11. Classifier performance on SiSt transition detection. It illustrates the performance on the SiSt and StSi transition part only for 0.4*s*n sliding window (n=5). Red lines represent SiSt, blue lines represent not Sit-to-Stand

Reporting Duration Error for True Positives is important because increasing the confidence threshold can raise precision though this comes at the cost of reduced recall. This metric, therefore, evaluates duration accuracy under high-confidence detection settings, while the Duration Error for All Detections reflects overall system performance when FP transitions were also considered. Table III reports the duration measurement results.

## IV. DISCUSSION

This study introduces a SiSt transition classification and duration measurement methodology utilizing multimodal data collected from the Smart Lacelock sensor. The sensor captures signals from a load cell, a three-axis accelerometer, and a three-axis gyroscope, providing a comprehensive dataset for activity recognition.

One of the primary challenges involved a significant class imbalance, with SiSt transition instances comprising less than one-tenth of the total dataset. To address this, we employed a combination of the TSAUG algorithm for signal-level augmentation and the Synthetic Minority Over-sampling Technique (SMOTE) for feature-level augmentation. This dual-augmentation strategy helped create a more balanced dataset, facilitating robust and generalizable classifier training.

Another challenge arose during the labeling process because of the considerable variation in SiSt transition durations across participants. To mitigate this, sliding window lengths of different sizes were systematically explored. The result in Fig. 9 demonstrates that increasing the sliding-window duration from 0.4 s (0.4*n, n = 1) to 2 s (0.4s*n, n = 5) substantially improved model performance for SiSt transition detection. The 2s sliding window achieved higher precision (0.81 vs. 0.75) and recall (0.77 vs. 0.69), resulting in an F1-score improvement from 0.72 to 0.8. This pattern held across other window sizes as well.

Moreover, Fig. 9 also shows that when the total sliding window length is close to 2 s, even with different primary window sizes (i.e., 0.6s*n, 0.7s*n, n=3), the F1 score remains close to its maximum. This is likely because a ~2 s window captures the full SiSt transition pattern for most participants. However, when the window size moves far from 2 s, performance decreases. Larger windows may include parts of the StSi transition, making it harder for the model to distinguish the activities and generalize well.

The load cell signal, exhibiting significant differences during SiSt and StSi transitions, further strengthened the classifier's discriminatory power. Apart from that, the load cell signal's distinct shape helped postprocessing and improved the results. Thus, combining load cell signal features with accelerometer and gyroscope signal features resulted in a classifier with the highest F1 score (0.8).

To further assess the contribution of the load cell, the system was evaluated without the inclusion of the load cell signal (Table II). In this configuration, performance declined notably, with the highest F1-score dropping from 0.8 (with load cell) to 0.62 (without load cell). Both precision and recall decreased across all window lengths, indicating that the load cell plays a critical role in distinguishing weight-shift patterns unique to the SiSt transition. The load cell captures the characteristic valley-to-peak-to-valley dynamics associated with pressure redistribution on the feet, which are not fully represented in the inertial data alone. Consequently, removing this modality weakens the separability between SiSt transitional and non-transitional activities, leading to an increased rate of misclassification.

These findings underscore the importance of multimodal sensor fusion in reliable SiSt detection. While IMU signals capture orientation and acceleration, the load cell provides complementary ground-reaction force information that anchors the temporal boundaries of the transition. The combined features thus yield both high temporal sensitivity and contextual precision, which are crucial for accurately identifying short-duration postural changes.

The duration measurement results followed the same trend as the classification performance, as observed in Table III. As classification improved with increasing sliding window size up to 2s, duration accuracy also improved.

For correctly detected transitions (True Positives), the lowest error was achieved with the 0.4 s*5 sliding window, resulting in an MAE of $0.047 \pm 0.07$s and an RMSE of $0.079 \pm 0.12$ s. This indicates very precise estimation of the start and end of the SiSt transition when detection was accurate. A similar pattern was observed when evaluating all detections, including missed transitions. The 0.4 s*5 sliding window again produced the lowest overall errors. Smaller than 2 s sliding windows likely lacked sufficient temporal context, while larger than 2 s sliding windows reduced temporal precision, leading to increased error.

Overall, the 2 s sliding window (0.4 s *n, n=5) provided the best balance between detection performance and duration measurement accuracy, confirming that proper window selection is critical for reliable estimation of SiSt transitions and durations.

However, a notable limitation is the need for precise synchronization of both left and right foot data, as misalignment could confuse the classifier. This synchronization requirement poses a practical challenge but is critical for achieving optimal classification results.

## V. CONCLUSION

This study presents a method for detecting SiSt transitions and measuring their duration in older adults using the novel Smart Lacelock sensor, a shoe-mounted device that combines a load cell and an IMU. Class imbalance was addressed using TSAUG and SMOTE, and different sliding window sizes were tested to account for variation in transition duration. The optimal configuration was 0.4 s × 5 (2 s total window). The Bagged Tree classifier attained an accuracy of 0.98 and an F1 score of 0.8. Including load cell data significantly improved detection performance (F1 score of 0.8) compared with using IMU data alone (F1 score of 0.62). For duration measurement, again the 2 s (0.4 s*n, n=5) sliding window achieved the lowest error, with an MAE of $0.047 \pm 0.07$s and an RMSE of $0.079 \pm 0.12$s for correctly detected transitions.

Overall, the results demonstrate that the novel Smart Lacelock sensor shows a promising outcome for continuous mobility monitoring and fall-risk assessment in older adults.

## ACKNOWLEDGMENT

Author Md R. Islam used ChatGPT to enhance this manuscript's clarity and language quality and to create graphical abstract figures. All scientific insights, analyses, and conclusions are the sole work of the authors.

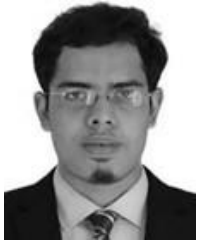

**Md Rafi Islam** received his B.Sc. in electrical and electronic engineering from Chittagong University of Engineering and Technology (CUET), Bangladesh, in 2017. He is currently pursuing a Ph.D. in electrical and computer engineering at The University of Alabama, Tuscaloosa, AL, USA.

He previously served as a Lecturer in the Department of Electrical and Electronic Engineering at Premier University, Chittagong. His research interests include embedded systems, machine learning, wearable robotics, and intent recognition.

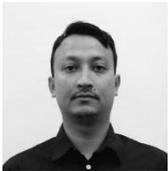

**Md Rejwanul Haque** is an R&D Engineer at Genesis Medtech USA Inc., specializing in the development of robotic surgical instruments and advanced medical devices. His work focuses on precision mechanism design, cable-actuated and articulated systems, and high-reliability assembly for minimally invasive applications. His research interests include surgical robotics, miniature actuation technologies, and system-level design optimization for enhanced clinical performance. He received his Ph.D. in Mechanical Engineering from the University of Alabama..

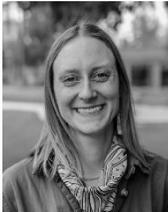

**Elizabeth "Lizzie" Choma** received her PhD in Nursing from the University of Minnesota and her Doctor of Physical Therapy degree from the University of Montana. She is a board-certified specialist in geriatric physical therapy and an Assistant Professor within Whitworth University's Doctor of Physical Therapy program. Lizzie's research interests include using implementation science to explore new ways to deliver exercise-based balance exercises to older adults in cardiac rehabilitation settings. Her doctoral research was supported by the Foundation for Physical Therapy Research.

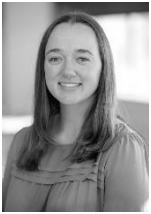

**Shannon Hayes** received her Bachelors of Science in Nursing from the University of Minnesota in 2022, graduating summa cum laude and receiving an additional minor degree in mathematics. Her professional experiences, both clinical and research related, encompass a variety of clinical settings with various patient diagnoses and complexities. She has made meaningful contributions to the understanding of health & exercise behaviors in older adults as well as the development of novel therapies for adult patients with hematologic malignancies and other blood disorders. Shannon currently resides in Durham, NC and aims to continue her work in clinical research.

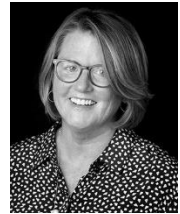

**Siobhan McMahon**, PhD, MPH, GNP-BC, is an Associate Professor at the University of Minnesota School of Nursing. As a nurse scientist, her research focuses on developing, testing, implementing and disseminating technology-enhanced strategiesMPH to reduce falls and promote physical activity among community-dwelling older adults. She earned her PhD in Nursing and Innovation from Arizona State University and her Master's degree from Marquette University. Dr. McMahon is also a board-certified gerontological nurse practitioner.

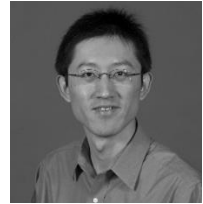

**Xiangrong Shen** received his Ph.D. in mechanical engineering from Vanderbilt University, Nashville, TN, USA, in 2006, followed by a two-year postdoctoral training in rehabilitation robotics at the same institution.

He is a Professor in the Department of Mechanical Engineering at The University of Alabama, Tuscaloosa, AL, USA, where he has been a Faculty Member since 2008. His research interests include assistive and rehabilitation robotics, focusing on robotic prostheses, power-assist orthoses, and therapeutic robotic platforms. His work has been funded by NSF and NIH, including an NSF CAREER Award in 2014.

Prof. Shen is the ME Dynamic Systems and Control (DSC) Group Chair and a Fellow of the Alabama Life Research Institute (ALRI). He previously served as an Associate Editor for Control Engineering Practice.

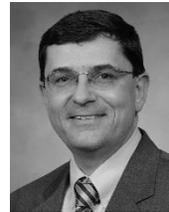

**Edward Sazonov** (M'02, SM'11) received his Diploma in systems engineering from Khabarovsk State University of Technology, Russia, in 1993, and a Ph.D. in computer engineering from West Virginia University, USA, in 2002.

He is a James R. Cudworth Endowed Professor in the Department of Electrical and Computer Engineering at The University of Alabama, Tuscaloosa, AL, USA. He heads the Computer Laboratory of Ambient and Wearable Systems. His work focuses on wearable sensors for monitoring behaviors such as food intake, physical activity, and smoking, with devices including the Automatic Ingestion Monitor (AIM), SmartStep, and the Personal Automatic Cigarette Tracker (PACT). His research interests span wearable devices, behavioral informatics, and biomedical signal processing.

Prof. Sazonov has received awards such as the President's Research Award at the University of Alabama and served as a Fulbright Distinguished Chair at the University of Newcastle, Australia. He currently serves as a Specialty Chief Editor for Wearable Electronics and an associate editor for several IEEE journals.